\documentclass[conference]{IEEEtran}
\IEEEoverridecommandlockouts
\usepackage{parskip}
\usepackage{framed,multirow}

\usepackage{amssymb}
\usepackage{latexsym}

\usepackage[table,xcdraw]{xcolor}
\usepackage{amsmath,amsfonts}
\usepackage[ruled,vlined]{algorithm2e}
\usepackage{url}

\usepackage{lipsum}
\usepackage{mathtools}
\usepackage{subcaption}
\usepackage{xurl}
\usepackage{array}
\usepackage{booktabs}
\usepackage{tabulary}
\usepackage{algorithmic}
\usepackage{graphicx}
\usepackage{textcomp}
\usepackage{color,soul}
\usepackage{float}
\usepackage{multirow}
\usepackage{hyperref}
\usepackage{xr}
\usepackage{cleveref}
\usepackage[utf8]{inputenc}

\DeclareUnicodeCharacter{2265}{\ensuremath{\geq}}
\DeclareUnicodeCharacter{2264}{\ensuremath{\leq}}

\definecolor{newcolor}{rgb}{.8,.349,.1}

\DeclareRobustCommand{\IEEEauthorrefmark}[1]{\smash{\textsuperscript{\footnotesize #1}}}

\def\BibTeX{{\rm B\kern-.05em{\sc i\kern-.025em b}\kern-.08em
    T\kern-.1667em\lower.7ex\hbox{E}\kern-.125emX}}

\begin{document}
\include{pythonlisting}

\title{Ugly Ducklings or Swans: A Tiered Quadruplet Network with Patient-Specific Mining for Improved Skin Lesion Classification }

\author{\IEEEauthorblockN{
Nathasha Naranpanawa\IEEEauthorrefmark{1}\textsuperscript{*}, 
H. Peter Soyer\IEEEauthorrefmark{2}, 
Adam Mothershaw\IEEEauthorrefmark{2}, 
Gayan K. Kulatilleke\IEEEauthorrefmark{1}, \\
Zongyuan Ge\IEEEauthorrefmark{3,}\IEEEauthorrefmark{4}, 
Brigid Betz-Stablein\IEEEauthorrefmark{2} and 
Shekhar S. Chandra\IEEEauthorrefmark{1}}

\IEEEauthorblockA{\IEEEauthorrefmark{1}School of Electrical Engineering and Computer Science,
The University of Queensland, Brisbane, Australia\\}
\IEEEauthorblockA{\IEEEauthorrefmark{2}Frazer Institute,
The University of Queensland, Dermatology Research Centre, Brisbane, Australia\\}
\IEEEauthorblockA{\IEEEauthorrefmark{3}Monash Medical AI,
Monash University, Australia\\}
\IEEEauthorblockA{\IEEEauthorrefmark{4}Department of Data Science and AI,
Faculty of IT, Monash University, Australia\\}
}

\maketitle
\begingroup\renewcommand\thefootnote{*}
\footnotetext{Corresponding author: nathasha.naranpanawa@uq.edu.au}
\endgroup
\begin{abstract}

An ugly duckling is an obviously different skin lesion from surrounding lesions of an individual, and the ugly duckling sign is a criterion used to aid in the diagnosis of cutaneous melanoma by differentiating between highly suspicious and benign lesions. However, the appearance of pigmented lesions, can change drastically from one patient to another, resulting in difficulties in visual separation of ugly ducklings. Hence, we propose DMT-Quadruplet - a deep metric learning network to learn lesion features at two tiers - patient-level and lesion-level. We introduce a patient-specific quadruplet mining approach together with a tiered quadruplet network, to drive the network to learn more contextual information both globally and locally between the two tiers. We further incorporate a dynamic margin within the patient-specific mining to allow more useful quadruplets to be mined within individuals. Comprehensive experiments show that our proposed method outperforms traditional classifiers, achieving 54\% higher sensitivity than a baseline ResNet18 CNN and 37\% higher than a naive triplet network in classifying ugly duckling lesions. Visualisation of the data manifold in the metric space further illustrates that DMT-Quadruplet is capable of classifying ugly duckling lesions in both patient-specific and patient-agnostic manner successfully.


\end{abstract}

\section{Introduction} \label{sec:intro}

\begin{figure*}[ht]
\centering
\includegraphics[scale=0.6]{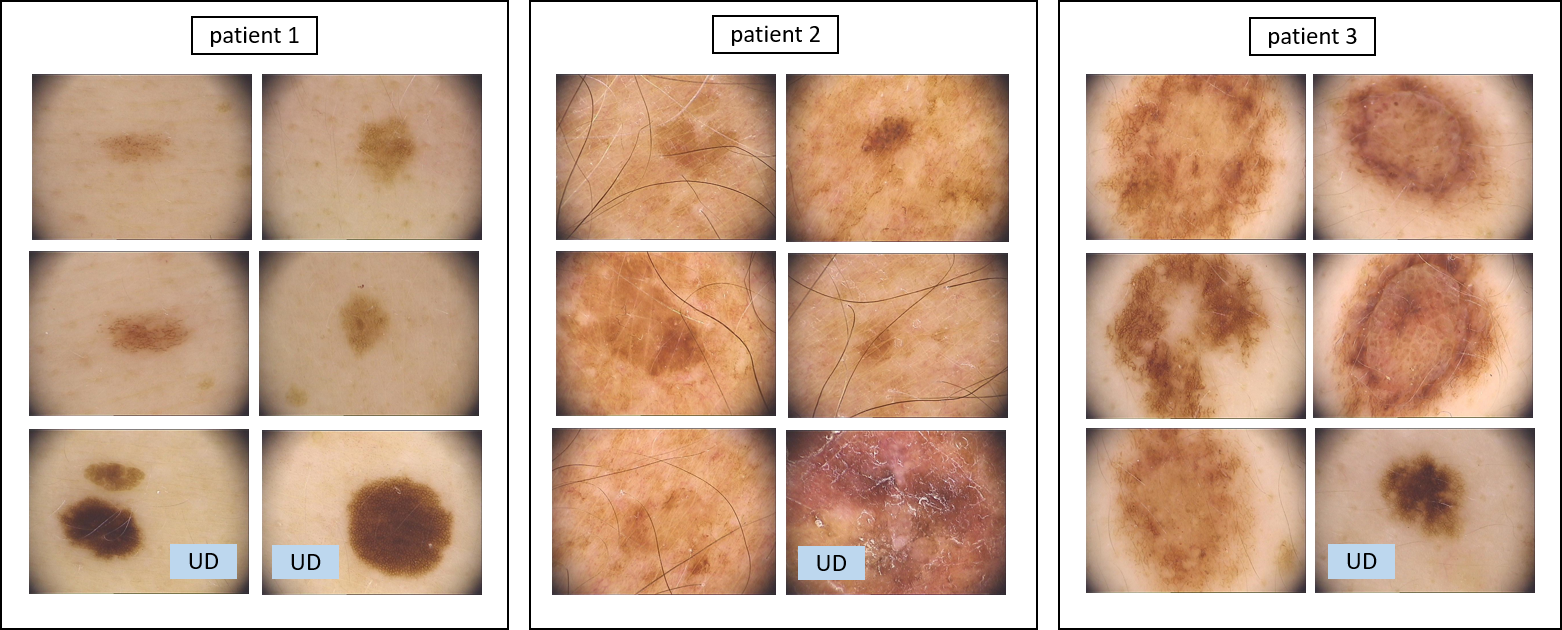}
\caption{Visual lesion characteristics vary among patients. Figure shows sets of lesions from 3 different patients. Each lesion set is from the same body-site of the particular patient, and ugly ducklings are marked with a `UD' label.}
\label{fig:1}
\end{figure*}

In clinical practice, the most common and established criteria for visually identifying a malignant melanoma is the ABCDE (Asymmetry, Border, Colour, Diameter, Evolution) criteria \cite{Rigel2005}. This criterion indicates that a skin lesion that lacks in symmetry (Asymmetry), has a spreading or an irregular edge (Border), a variegated colour (Colour), diameter is larger than 6mm (Diameter), and is changing in size and colour over time (Evolution) might be a melanoma. Nonetheless, there exists cases where malignant melanomas do not conform to this criteria, requiring an alternate recognition strategy. Therefore, \cite{JJ1998a} introduced the concept of an 'Ugly Duckling' as an additional criteria for visual inspections of the skin for melanoma. This criterion was developed based on the fact that naevi on an individual tend to resemble one another, whereas malignant melanoma often deviates from this nevus pattern and stands out from its peers on a common body region. Hence, an 'Ugly Duckling' lesion is defined as an visually different nevus from its surrounding naevi on an individual, and is considered suspicious for malignancy. 

When identifying an ugly duckling among other lesions, clinicians have access to contextual information as they can observe all lesions on a patient's skin and make a judgement on the visual similarity of the lesions. This similarity judgement is an important cognitive process in humans. By comparing perceptual representations, they are able to perform tasks such as recognition and categorization. This concept of similarity underlies most machine learning techniques. Thus, we can use the same notion of perception to apply a deep learning-based solution to recognize ugly ducklings for the detection of malignant melanoma. 

However, disparities among individuals and their respective set of lesions can exist in terms of colour, shape, size and distribution. The characteristics of lesions of one individual might be completely different to that of another \cite{Wazaefi2013}. This inconsistency is further indicated by the considerable inter-observer variability in selecting ugly duckling lesion, as expert physicians also differ in clinical experience and visual perception \cite{Gaudy-Marqueste2017, Scope2008a}. When it comes to implementing AI-based ugly duckling recognition methods, disregarding this pattern variability between individuals and training a model only at lesion-level might be disadvantageous, leading to incorrect predictions. For an example, a naive classifier trained with individual lesion images of several different individuals will not identify patient-specific lesion similarities, as an ugly duckling on one patient might look completely normal on another patient's skin (Figure \ref{fig:1}). However, the ontology of a skin lesion dataset is such that the sub-classes (normal and ugly duckling lesions) are not mutually exclusive among individuals although intra-patient features might be different. Thus, the classification of ugly duckling lesions pose a unique and challenging task of taking into account both inter-patient (patient-level) and intra-patient (lesion-level) feature representations. 

We propose to solve this problem by using metric learning - specifically a triplet network with patient-specific sample mining and dynamic separating margins, which is then extended to a tiered quadruplet network that is capable of remaining patient-agnostic while learning patient-specific representations.

Metric learning infers contextual information automatically from data to measure similarity, which it will separate naevi displaying the ugly duckling sign from the common moles or non-malignant naevi of an individual. With this system, given a set of images of naevi from a patient, the ugly ducklings can be detected easily and further observed to decide malignancy.

In summary, the main contributions of this work are as follows:

\begin{enumerate}
    \item We propose a novel patient-specific metric-learning method for improved classification of ugly duckling lesions in largely imbalanced skin lesion datasets.
    \item We propose a new patient-specific triplet mining strategy that is capable of capturing the patient-level differences accurately.
    \item We show that extending this triplet sampling method into a quadruplet sampling method with patient-specific dynamic margins is capable of bypassing the limitations of naive sample mining by incorporating more contextual information from two tiers of separation while maintaining the semantic differences.
    \item Using the learning metric for similarity, we show that classification of ugly duckling lesions of a patient is improved compared to the traditional classification methods.
    \item We build a novel deep-learning based pipeline that can be used to automatically find ugly ducklings on a patient's skin by providing all lesion images of that patient to the network. %finetune

\end{enumerate}

The rest of the paper is organized as follows: Section \ref{sec:related work} provides a literature review on classification and metric learning methods applied for suspicious naevi classification. Section \ref{sec:methods} presents the proposed method, and Section \ref{sec:experiments} details the experimental evaluations. This is followed by Section \ref{sec:results} where results are presented and their implications are discussed.

\section{Related work} \label{sec:related work}

With the recent advances in deep learning, many domains of medical image analysis, including dermatology, have seen promising results. Given the advantage of having public skin lesion datasets such as the The International Skin Imaging Collaboration (ISIC) archive \cite{InternationalSkinImagingCollaboration2016a} available for analysis, there has been many work that explored deep learning methods for melanoma detection and skin classification over the past few years. Some comprehensive reviews on the topic can be found in \cite{Dildar2021}, \cite{Hasan2023}, \cite{Liu2023}, \cite{Yaqoob2023}, \cite{Wu2022a}, \cite{Bhatt2022} and \cite{Cassidy2022}. However, we focus on the classification of ugly duckling lesions in our work, for which public annotations are currently unavailable. 

\subsection{Ugly duckling recognition}

% \hl{add skin lesion classification work - malignant vs benign, cite review papers
% melanoma in general - review paper
% we're focusing on ugly ducklings}

%metric learning for ugly duckling classification
% With the recent advances in deep learning, many domains of medical image analysis, including dermatology, have seen promising results. 
Despite the advances with melanoma analysis, only a few previous work have focused on improving the identification of cutaneous ugly duckling lesions using deep learning methods.

%Wide-field images 
The use of wide-field images from total-body photography are popular among the few Computer-Aided Detection (CAD) works that have focused on detecting the ugly duckling sign \cite{Birkenfeld2020a, Mohseni2021, Soenksen2021a, Useini2021, Garcia2022}. Wide-field images are an advantage in ugly duckling related work as by definition, an ugly duckling is a contextual observation in comparison to neighbouring lesions on a body-site, and wide-field images provide a way to include this contextual information. 

The common workflow in these proposed methods is to first extract the lesions out of the wide-field images, and train the detectors with them. Labelling of the lesions might be performed before or after extraction. With the use of wide-field images, even straightforward methods such as logistic regression models show promising results \cite{Birkenfeld2020a}. Modelling the ugly duckling detection as an outlier detection on wide-field is also popular, with results demonstrating that patient-level analysis of wide-field images is advantageous. As proposed by the work of \cite{Mohseni2021} and \cite{Garcia2022}, autoencoders trained with the extracted lesions can be used successfully for ugly duckling outlier detection. According to the study by \cite{Soenksen2021a}, deep convolution neural networks with transfer learning also show promising results for identifying inter-lesion dependencies for the classification of ugly ducklings. Further indicating that well-optimized deep learning methods can be efficiently used for accurate assessment of suspicious pigmented lesions, \cite{Yu2021} analysed individual lesion images instead of wide-field images to identify ugly ducklings. They learn a patient-specific contextual embedding by modelling the dependencies among lesions using a Transformer encoder, which are then used to perform patient-level and lesion-level predictions concurrently. Similarly, \cite{AlZegair2022} shows that feature extraction of suspicious lesions with variational autoencoders followed by random forest and artificial neural network classifiers can obtain high classification accuracy. All of these works demonstrate that patient-level analysis or incorporation of patient-specific contextual information is more advantageous for identifying ugly duckling lesions accurately.

Most of the above studies developed a ranking system to separate ugly ducklings from normal lesions \cite{Birkenfeld2020a, Mohseni2021, Soenksen2021a, Garcia2022}. A feature vector or an embedding for each lesion were extracted from the trained models, and the distance between these embeddings were used to determine a separating threshold or a ranking score for ugly ducklings. However, some studies pooled all of the extracted lesions from wide-field images together to train the models, resulting in patient-level contextual information being lost. Some thresholds were set based on clearly defined melanomas only \cite{Garcia2022}. These actions might have resulted in a more lesion-level threshold of separation than patient-level, leading to misidentification of ugly ducklings that had features not clearly distinguishable during inference with the models.  

Furthermore, based on the definition of an ugly duckling, datasets used for these studies would need specific ugly duckling annotations. While some studies labelled the ugly ducklings in their datasets with the help of board-certified dermatologists, a few others simply used existing labels of malignant melanoma in place of ugly ducklings. However, malignant melanoma annotations cannot be used interchangeably with ugly duckling labels given that not all melanomas present ugly duckling characteristics \cite{Rotemberg2021}. Thus, studies using malignant melanoma labels might have resulted in misclassification of actual ugly duckling lesions. Therefore, expert supervision is necessary to either annotate datasets correctly with ugly duckling labels, or evaluate the ranked lesions to test high agreement. This might be costly as time and expertise supervision is required. In addition, there are currently no public datasets that are specifically annotated with ugly duckling labels. These might be the reason that not many deep learning based methods have not been explored for ugly duckling recognition.

\begin{figure*}[t]
\centerline{\includegraphics[width=1\textwidth, scale=0.3]{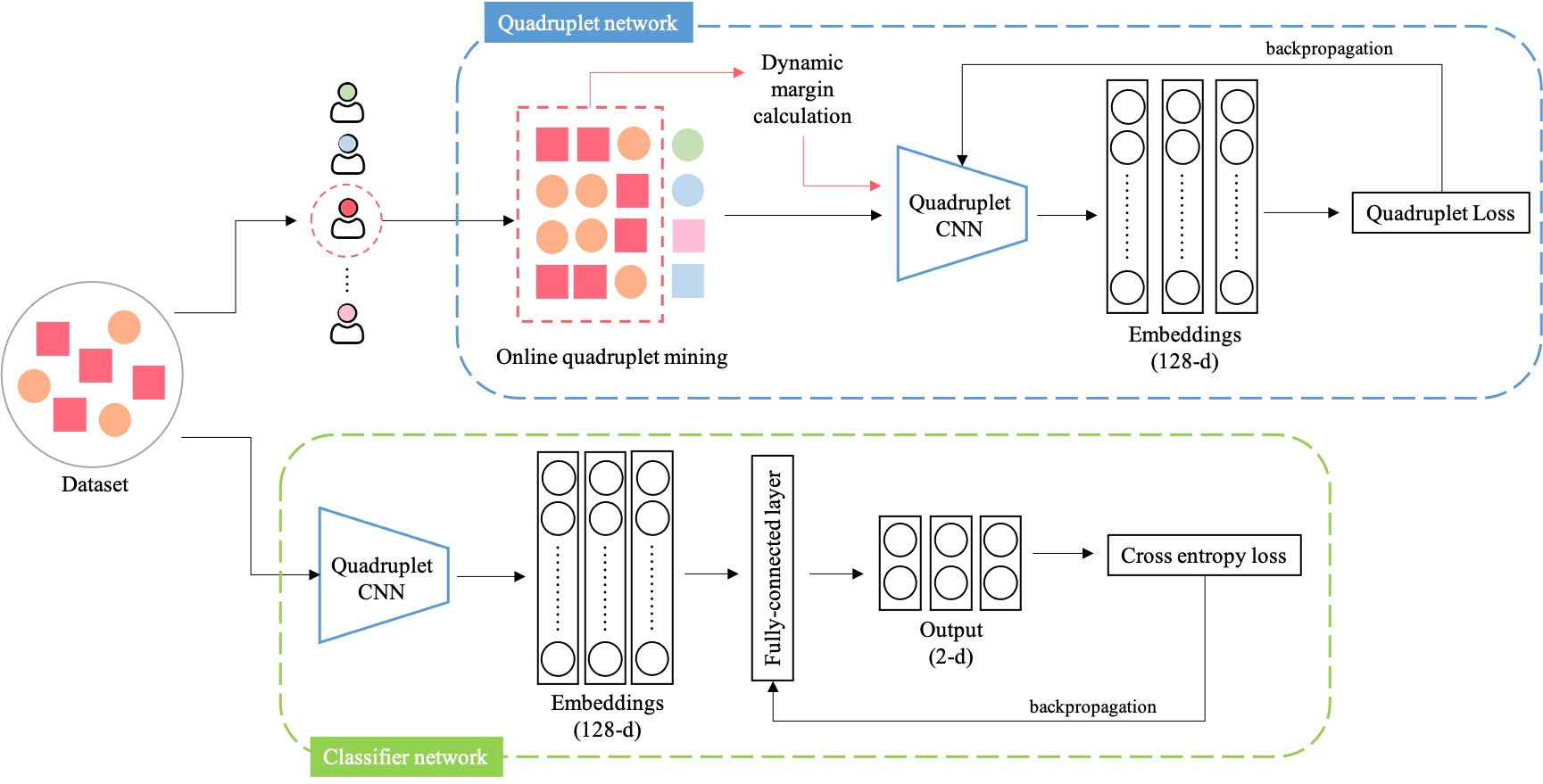}}
\caption{The proposed DMT-Quadruplet architecture. In the first stage, a Quadruplet network is trained with online patient-specific quadruplet mining and a dynamic margin. The trained quadruplet network is then used as a backbone feature extractor to train a simple CNN classifier in the second stage. During the testing phase, only the trained classifier network from the second stage is used for generating a lower dimensional embedding for each test image and performing binary classification.}
\label{fig:3}
\end{figure*}

\subsection{Metric learning for skin lesion classification}

In current literature, no metric learning approaches have been employed for the identification of ugly duckling lesions specifically. However, a few works on melanoma detection and skin lesion segmentation have explored the use of metric learning. 

As demonstrated by past work, traditional triplet loss is capable of achieving higher intra-class compactness and stronger inter-class separability in skin lesion and skin disease classification \cite{Ahmad2020, Sun2022}. A few studies have also improved skin lesion segmentation by using triplet networks to learn pixel embedding in the metric space \cite{Liu2019c, Cao2022}. In addition, \cite{Allegretti2020} proposed a trained triplet network to gather similar images from publicly available skin lesion datasets to support skin lesion diagnosis with content-based image retrieval.

Apart from that, most of the works using metric learning for skin lesion classification are focused on solving the class-imbalance problem in the respective skin lesion datasets used \cite{Hsu2022}. In cases of heavy imbalance, traditional classifiers might maintain a high overall accuracy because of the bias introduced from the large number of samples from the major class. Thus, complex balancing techniques have to be introduced within traditional classifiers to improve prediction accuracy on minority class \cite{Mazurowski2008a,Haixiang2017}. Contrastive learning, especially with triplet networks and the traditional triplet loss itself, is capable of addressing this class imbalance by generating a balanced contribution by each class in mined triplets. However, samples within the same class might present large visual differences, causing the feature distribution of that class to spread over a large space. By employing a class-center involved triplet loss, this distribution can be made more compact in the learned feature space as shown by previous work \cite{Lei2020, Ozturk2022, Chen2023}. Using a pretrained triplet network capable of generating the embeddings for the samples, the class centers can be calculated, which are then used in place of positive and negative samples within the mined triplet. This approach ensures the representations of samples from the same class are driven closer towards the class center. \cite{Ozturk2022} extends the center-oriented triplet loss with an adaptive margin value, where the margin value is automatically adjusted given cluster separation instead of training with a fixed value.

Despite these advances, no deep metric learning models have enforced a patient-specific separation for skin lesion classification focusing specifically on ugly duckling identification. Therefore, to the best of our knowledge, this is the first work employing a tiered and patient-specific triplet network for ugly duckling lesion classification.

\section{Methods} \label{sec:methods}

\subsection{Overview of the proposed architecture}

% just a few sentences about the overall pipeline end-to-end

% Class imbalance is a challenging issue in the application of classification .  Triplet networks have been been able to solve this issue as they increase the contribution of minority class samples to the loss by creating multiple triplet instances including these samples \cite{Lei2020, Chen2023, Ozturk2022}.

Publicly available skin lesion datasets suffer from high imbalances as melanoma incidences are rare \cite{Combalia2019, Codella2018b, Tschandl2018a}. In order to accurately classify both normal and rare ugly duckling lesions, while accounting for both local and global feature representations, we propose a two-stage classification system. The overview of the proposed framework is presented in Figure \ref{fig:3}. The first stage involves training a quadruplet network, and the second stage utilizes the trained quadruplet network to generate a latent representation of samples in metric space and train a classifier on them.

\subsubsection{Classical triplet loss and triplet mining}

To better describe our method, we first introduce the functionality of a classical triplet network. 

A triplet network is trained on multiple triplets sampled from a dataset, where each triplet contains an anchor sample (\textit{a}), a positive sample (\textit{p}) from the same class as the anchor, and a negative sample (\textit{n}) from the opposite class. The traditional triplet loss then aims to minimize the $a-p$ distance, while maximizing the $a-n$ distance by at least a margin $\alpha$:

\begin{equation} \label{eq:1}
\scalebox{0.9}{$
\begin{split}
L_{triplet} = \frac{1}{N}\sum_{i=1}^{N}[ d(a^{i},p^{i}) - d(a^{i},n^{i}) + \alpha  ]]
\end{split}$}
\end{equation}

where $d$ is the Euclidean distance between a lower dimensional embedding of two samples. For each training iteration, a batch of N samples is used to select random triplets. The triplet loss calculated per triplet instance is then averaged within that batch.

% \hl{might have to replace the use of N above as it confuses with N in pseudo code}

The selection of triplets, which is referred to as 'triplet mining', can be performed either offline or online. In the offline triplet mining strategy, we form the triplets at the beginning of each epoch and feed batches of those triplets throughout the epoch. However, this is inefficient as the number of triplets that need to be computed are high and as not all of them contribute to the learning of the network. This can be alleviated by the online mining strategy where useful triplets are mined on the fly for a mini-batch of samples.

However, not all online triplets are `valid' as some of them don't have two positives and one negative sample. To find out the valid and useful triplets, strategies such as semi-hard mining and hard mining can be used. In hard-mining, the negative sample is closer to the anchor than the positive and the loss is positive as well as greater than $\alpha$. In semi-hard mining, the negative sample is further than the positive but not larger than the margin, and the loss is positive.

\subsubsection{Triplet Loss with patient-specific mining approach}

In the first stage of our proposed method, we first implement a patient-specific triplet loss to train a feature extractor capable of capturing discriminate inter-patient features in samples. The triplet network consists of a CNN backbone, whose output is a 128-dimensional feature vector. The distances for the triplet loss are calculated between corresponding embeddings within a mined triplet to update the CNN weights.  

\begin{figure}[t]
\centerline{\includegraphics[width=0.5\textwidth, scale=0.5]{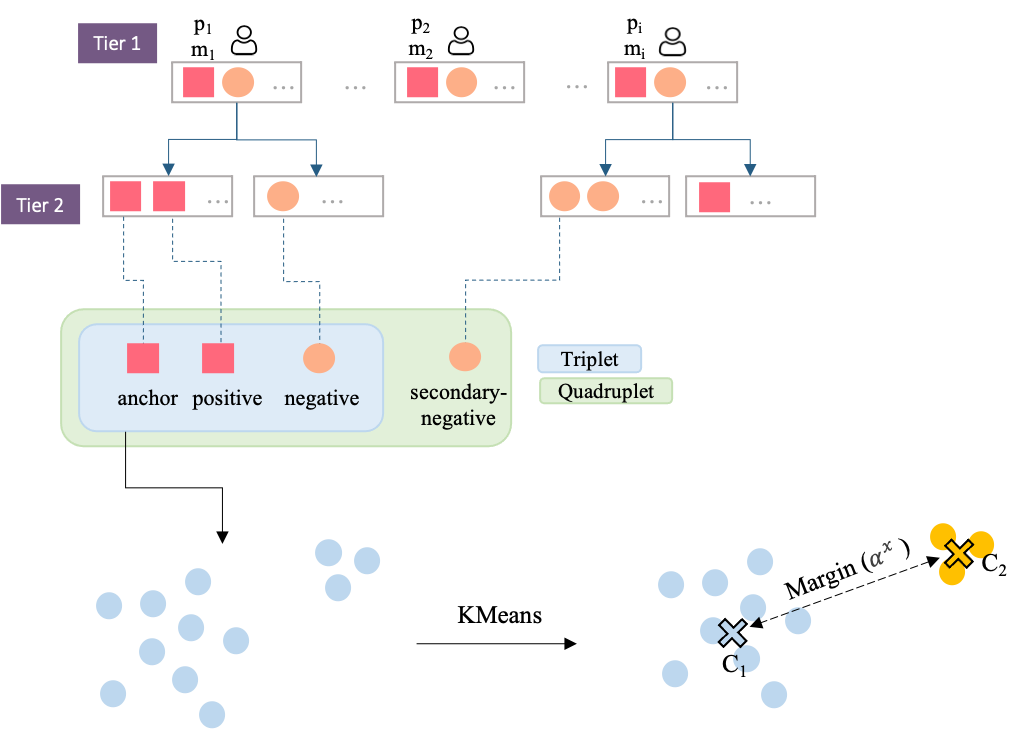}}
\caption{Patient-specific mining approach. The blue boxes indicate the mining of a triplet within a patient, where anchor and positive are from the same class, while negative is from the opposite class. For the Tiered Quadruplet loss, we mine a fourth sample (secondary-negative) as indicated by the green box. The secondary-negative can be of either class, but from a \textit{different} patient. For the Tiered Quadruplet with a Dynamic Margin, we calculate an online margin of separation for two distinct clusters of embeddings of an individual using KMeans.}
\label{fig:2}
\end{figure}

In a naive triplet mining approach, the patient-level separation is disregarded, instead mining the triplets based only on their class labels. However, we consider the fact that there are two levels of contextual information included in the particular problem of classifying skin lesions - lesion-level (intra-patient) and patient-level (inter-patient) differences. Therefore, we follow a patient-specific mining approach as illustrated in Figure \ref{fig:2}.

\begin{algorithm}[t]
\SetAlgoLined
\textbf{Input: } Training data $D$, number of epochs $e$, triplet margin $\alpha$, network hyperparameters \\
\textbf{Initialization: } $t, X, k$ \\
% \KwResult{Write here the result }
 
 \While{$t$ $<$ $e$}{
    list triplets $T$ \\
    Randomly select $k$ samples from $X$ individuals for iteration $t$ \\

    \For{ $x$ in $X$}{
        Mine all anchor-positive pairs $(A^{x},P^{x})$\\
        \For{$(a^{x},p^{x})$ in $(A^{x},P^{x})$}{
            Mine all negative samples $N^{x}$ from individual $x$ \\

            Create all triplets $(a^{x},p^{x}, n^{x})$ for $n^{x}$ in $N^{x}$ \\
            \For{each triplet $(a^{x},p^{x}, n^{x})_{i}$}{
                Generate embeddings $(w^{x}_{a},w^{x}_{p}, w^{x}_{n})_{i}$ with current model parameters\\
                % Compute $d((w^{x}_{a},w^{x}_{p})_{i})$ and $d((w^{x}_{a},w^{x}_{n})_{i})$ \\
                Calculate triplet loss $tl_{i}$ for each embedding triplet $(w^{x}_{a},w^{x}_{p}, w^{x}_{n})_{i}$  as in Equation \ref{eq:1} \\
                
            }
            From triplets with $tl > 0$, randomly select triplet $(a^{x},p^{x}, n^{x})$\\
            Add selected $(a^{x},p^{x}, n^{x})$ to $T$
        }
    }

    Calculate triplet loss for iteration $tl_{t}$ with all triplets in $T$ as in Equation \ref{eq:1} \\
    Backpropagate to update CNN parameters \\
 }
 \caption{Pseudo-code for patient-specific triplet loss} \label{alg:1}
\end{algorithm}

The pseudo code in Algorithm \ref{alg:1} outlines the patient-specific triplet mining and training procedure of the triplet network. We start by using a custom batch-sampler to create mini-batches of samples. For this, we randomly select an $X$ number of individuals from the training dataset, and randomly sample $k$ images from each individual between both classes of lesions. This ensures each mini-batch would see sufficient information of each individual's lesions.

Then, during each epoch $t$ of training, we consider only the samples belonging to a particular individual ($x$) to select the triplets for iteration.  To create the triplets, we first mine all pairs of anchor ($a^{x}$) and positive ($p^{x}$) samples. Then, for each anchor-positive pair, we create all possible triplet instances by considering each negative sample ($n^{x}$) of that individual in the mini-batch.  

For all such triplet instances created for an anchor-positive pair, a 128-dimensional embedding is then calculated using the most recently updated neural network backbone of the triplet network. For each embedding triplet, the triplet loss is computed using Equation \ref{eq:1} where the distance measure is the Euclidean distance between the embeddings. Based on the loss value of each triplet with the mini-batch, we then perform random-hard triplet mining to select the hardest negative sample corresponding to each anchor-positive pair. This procedure is repeated for all individuals within that mini batch to create a batch of useful triplets, effectively incorporating more patient-specific local information into the triplet. Finally, the backbone CNN parameters are updated according to the triplet loss calculated on the mined triplets. This patient-specific mining procedure ensures that the CNN will focus on learning better patient-level feature representations.

\subsubsection{Tiered quadruplet loss}

We further extend our patient-specific triplet loss into a Tiered Quadruplet (T-Quad) loss to incorporate more lesion-level global context in the training of the triplet. With this, our aim is to identify the similarities between the lesions of individuals, as well as dissimilarities between individuals in the data. We model this as a two-tier problem where Tier 1 is composed of different individuals in the dataset, while Tier 2 is the collection of lesions of each individual (Figure \ref{fig:2}). 

The Tiered Quadruplet network follows the same sample mining procedure as in Algorithm \ref{alg:1}, but now we mine a fourth sample for each $(a^{x},p^{x})$ pair. We refer to this additional sample as a secondary-negative ($sn$) which is mined from a different patient ($y$) from the mini-batch. This secondary-negative can be of the same class or opposite class as the anchor ($a^{x}$). Similar to above, the hardest secondary-negative sample is chosen randomly for each $(a^{x},p^{x})$ pair. Thus, the Tiered Quadruplet loss takes in quadruplets of samples in the form $(a^{x},p^{x},n^{x},sn^{y})$ for each batch, and the loss is computed as:

\begin{align} \label{eq:2}
\begin{split}
&L_{patient\:level^{i}} = d(a^{i},p^{i}) - d(a^{i},n^{i}) + \alpha \\
&L_{lesion\:level^{i}} = d(a^{i},p^{i}) - d(a^{i}, \color{blue} sn^{i} \color{black}) + \color{blue} \beta \color{black}\\
&L_{t-quad} = \frac{1}{N}\sum_{i=1}^{N}[L_{lesion\:level^{i}} + L_{patient\:level^{i}}] \\
\end{split}
\end{align}

where $\beta$ ($> \alpha$) is a coarse margin of patient separation. Thus, with the tiered loss, the network learns a more global representation of lesions as Tier 2 classes (normal and UD lesions) are mutual across Tier 1 classes (individuals).

\subsubsection{Dynamic margin involved tiered quadruplet loss}

In naive triplet mining strategies, the margin of separation $\alpha$ is set to a default value across all mini-batches of training. However, we argue that this margin would be different from one individual to another based on the previously mentioned observation of their phenotypic features being different among themselves. Therefore, based on our assumption that the metric learning benefits from patient-specific information to learn a better separation of lesion characteristics, we further extend our Tiered Quadruplet loss to include a patient-specific dynamic margin instead of a global fixed margin.

The pseudo code in Algorithm \ref{alg:2} outlines the tiered quadruplet mining and calculation of the patient-specific dynamic margin for training the quadruplet network. For each mini-batch of embeddings during training, we calculate a dynamic margin of separation between two drastically different clusters of embeddings of a patient (Figure \ref{fig:2}). For each patient in the mini-batch, the embeddings are clustered unsupervised into two groups using the K-Means algorithm, and the distance between the two cluster centroids is used as the dynamic margin for mining $(a^{x},p^{x},n^{x})$ triplets from that particular patient for the considered mini-batch. Each patient-specific dynamic margin $\alpha^{x}$ is used to compute the Dynamic Margin Tiered Quadruplet (DMT-Quad) loss at each iteration as:

\begin{align} \label{eq:3}
\begin{split}
&L_{patient\:level^{i}} = d(a^{i},p^{i}) - d(a^{i},n^{i}) + \color{blue}
\bf{\alpha^{x}} \color{black} \\
&L_{lesion\:level^{i}} = d(a^{i},p^{i}) - d(a^{i},sn^{i}) + \beta \\
&L_{dmt-quad} = \frac{1}{N}\sum_{i=1}^{N}[L_{lesion\:level^{i}} + L_{patient\:level^{i}}] \\
\end{split}
\end{align}

For mining the $sn^{y}$ from a different patient, a fixed default margin $\beta$ is used. This is because the calculation of dynamic margins between patients would make the quadruplet mining process excessively computationally heavy.

\begin{algorithm}
\SetAlgoLined
\textbf{Input: } Training data $D$, number of epochs $e$, patient-level margin $\beta$, network hyperparameters \\
\textbf{Initialization: } $t, X, k$ \\
% \KwResult{Write here the result }
 \While{$t$ $<$ $e$}{
    list quadruplets $Q$ \\
    Randomly select $k$ samples from $X$ individuals for iteration $t$ \\
    \For{ $x$ in $X$}{
        \color{blue}
        \textbf{Find dynamic margin for all embeddings of individual }$x$\\
        \color{black}
        Generate embeddings for all samples of $x$ \\
        Apply KMeans on all embeddings for clustering into 2 classes \\
        Calculate the distance between the two cluster centroids as $\alpha_{x}$\\
        % \texttt{\\} \\
        Mine all anchor-positive pairs $(A^{x},P^{x})$\\
        \For{$(a^{x},p^{x})$ in $(A^{x},P^{x})$}{
            \color{blue}
            \textbf{Calculate lesion-level loss}\\
            \color{black}
            Mine all negative samples $N^{x}$ from individual $x$ \\

            Create all triplets $(a^{x},p^{x}, n^{x})$ for $n^{x}$ in $N^{x}$ \\
            \For{each triplet $(a^{x},p^{x}, n^{x})_{i}$}{
                Generate embeddings $(w^{x}_{a},w^{x}_{p}, w^{x}_{n})_{i}$ with current model parameters\\
                % Compute $d((w^{x}_{a},w^{x}_{p})_{i})$ and $d((w^{x}_{a},w^{x}_{n})_{i})$ \\
                Calculate triplet loss $tl_{i}$ for each embedding triplet $(w^{x}_{a},w^{x}_{p}, w^{x}_{n})_{i}$  as in Equation \ref{eq:3}  using dynamic margin $\alpha_{x}$ \\      
            }
            From the subset of triplets where $tl > 0$, randomly select triplet $(a^{x},p^{x}, n^{x})$\\
            \color{blue}
            \textbf{Calculate patient-level loss}\\
            \color{black}
            Mine all secondary-negative samples $SN^{y}$ where $y$ != $x$ \\
            Create all triplets $(a^{x},p^{x}, sn^{y})$ for $sn^{y}$ in $SN^{y}$ \\
            \For{each triplet $(a^{x},p^{x}, sn^{y})_{j}$}{
                Generate embeddings $(w^{x}_{a},w^{x}_{p}, w^{y}_{sn})_{j}$ with current model parameters\\
                % Compute $d((w^{x}_{a},w^{x}_{p})_{i})$ and $d((w^{x}_{a},w^{x}_{n})_{i})$ \\
                Calculate triplet loss $tl_{j}$ for each embedding triplet $(w^{x}_{a},w^{x}_{p}, w^{y}_{sn})_{j}$  as in Equation \ref{eq:3} using patient-level margin $\beta$
            }
            From the subset of triplets where $tl > 0$, randomly select triplet $(a^{x},p^{x}, sn^{y})$\\
            Add selected $(a^{x},p^{x},n^{x},sn^{y})$ to $Q$
        }
    }
    Calculate tiered quadruplet loss for iteration $tl_{t}$ with all triplets in $Q$ as in Equation \ref{eq:3}  with $\beta$ and corresponding $\alpha_{x}$ \\
    Backpropagate to update CNN parameters
 }
 \caption{Pseudo-code for DMT-Quadruplet loss} \label{alg:2}
\end{algorithm}

\subsubsection{Classification Network}
In the second stage of our proposed method, we train a CNN classifier to be able to identify normal and ugly duckling lesions. As illutsrated in Figure \ref{fig:3}, the trained quadruplet network acts as the feature extractor of the CNN classifier, where a fully-connected layer is added in the end to produce a 2-dimensional output that can be used for binary classification. All layers of the quadruplet network are frozen, and only the parameters of the last layer are updated in the classifier.

\begin{table*}[t]
\caption{Classification performance of triplet-based classifiers on the test set}\label{tab:table1}
\resizebox{\textwidth}{!}{%
\begin{tabular}{c|l|ccccccc|}
\cline{2-9}
\multicolumn{1}{l|}{} & \textbf{Method} & \textbf{Specificity} & \textbf{Sensitivity} & \textbf{Recall} & \textbf{Precision} & \textbf{F1-score} & \textbf{AUC} & \textbf{Accuracy} \\ \hline
\multicolumn{1}{|c|}{\multirow{4}{*}{\textbf{Baseline CNN classifiers}}} &ResNet18 & 96.4±1.3 & 46.2±6.2 & 71.3±2.8 & 66.3±3.0 & 68.1±2.6 & 84.0±0.0 & 94.0±1.4 \\
\multicolumn{1}{|c|}{} & ResNet34 & 96.3±0.6 & 47.4±4.4 & 71.9±2.3 & 66.1±2.7 & 68.4±2.4 & 88.7±3.1 & 94.0±0.8 \\
\multicolumn{1}{|c|}{} & VGG16 & 94.9±0.9 & 54.6±10.0 & 74.7±4.5 & 63.9±0.2 & 67.4±1.0 & \textbf{90.7±1.4} & 92.7±0.5 \\
\multicolumn{1}{|c|}{} & EfficientNeB0 & \textbf{97.6±0.2} & 40.4±0.5 & 69.0±0.2 & \textbf{68.9±0.8} & \textbf{68.9±0.3} & 89.7±1.8 & \textbf{95.0±0.0} \\ \hline
\multicolumn{1}{|c|}{\multirow{2}{*}{\textbf{Traditional metric losses}}} & Naïve Siamese & 93.4±0.3 & 60.1±2.5 & 76.8±1.1 & 62.5±0.3 & 66.3±0.4 & 86.8±1.4 & 91.7±0.5 \\
\multicolumn{1}{|c|}{} & Naïve Triplet & 93.9±1.6 & 51.9±2.6 & 76.4±3.5 & 63.3±1.0 & 66.9±0.6 & 88.5±0.0 & 92.0±1.4 \\ \hline
\multicolumn{1}{|c|}{\multirow{3}{*}{\textbf{Patient-specific metric losses}}} & Patient-specific Triplet (PS-Triplet) & 94.5±3.1 & 58.8±6.1 & 73.9±4.2 & 64.5±3.6 & 66.8±2.4 & 87.1±0.0 & 92.33±2.5 \\
\multicolumn{1}{|c|}{} & Patient-specific   Tiered Quadruplet (T-Quad) & 93.5±1.5 & 63.0±7.2 & 78.3±2.8 & 63.3±0.9 & 67.2±0.9 & 88.3±0.0 & 91.7±1.2 \\
\multicolumn{1}{|c|}{} & Patient-specific   Tiered Quadruplet + Dynamic Margin (DMT-Quad) & 91.7±1.4 & \textbf{71.2±5.9} & \textbf{81.4±2.2} & 62.2±0.8 & 66.3±0.9 & 90.2±0.0 & 90.3±1.3 \\ \hline
\end{tabular}
}
\end{table*}

\section{Experimental Evaluation} \label{sec:experiments}

\subsection{Data preprocessing}

The deidentified dataset used in this study was provided by the Dermatology Research Centre located within the Frazer Institute, The University of Queensland. It contains dermoscopy images of skin lesions from 59 participants that were originally collected under the Brisbane Naevus Morphology Study (BNMS) \cite{Sturm2014}. The data was collected from participants at multiple visits over a period of 3 years from October 2009 to October 2012. For each participant, all naevi ($\geq2mm$ and $\geq5mm$) at 16 body sites (excluding scalp, buttocks, mucosal surfaces, and genitalia of both sexes) had dermoscopic images taken with the FotoFinder® (Bad Birnbach, Germany) imaging system.

For the purpose of this study, all lesions were annotated as normal or ugly duckling by a board-certified dermatologist (HPS) with 30+ years of experience using an in-house annotation software. The software presents all lesions recorded per visit for a selected individual on the screen, allowing the annotator to observe and compare the lesions with each other rather than observe a single lesion at a time. Using this software, ugly duckling labels were assigned to lesions of each visit based on visually contrasting characteristics.   

To reduce the effect of large data imbalances, we discarded the records from participants who had no ugly ducklings recorded in any visit. For participants with ugly ducklings in some visits only, we also discarded the visits which had no ugly ducklings recorded. If a lesion of a participant had been labelled as UD at any visit, all lesions recorded from that visit were retained. If a lesion was labeled UD at one visit the labelling was kept consistent across all retained visits to reduce the confusion in feature learning for the neural network with inconsistencies. For normal lesions, all versions recorded across the retained visits were kept. The final dataset had a total of 10,493 images from 37 participants, with 10,174 normal lesions and 319 UDs. Thus, the ratio between UD to normal lesions is approximately 1:32. Of the final dataset, 21 ($\sim$57\%) paricipants were male and all participants were of Caucasian heritage. This dataset derived from the BNMS data will be referred to as the SkinUD dataset hereafter.

%6196 images from 57 participants, where the ratio of UD to normal lesions was approximately 1:108. 
The SkinUD dataset was split into training, validation, and test sets by patients in order to avoid data leakage. The images were originally 768$\times$576 pixels in size. During training, images are first center-cropped to 512$\times$512 pixels, which are then resized to 96$\times$96 pixels. Further, random horizontal and vertical flipping are applied as data augmentations. In addition, to alleviate the effect of the major class imbalance, we manually oversample UD lesions in the training set by duplicating the images by 10-fold.

In addition to the SkinUD dataset, we train and evaluate the performance of our proposed method on the ISIC2020 dataset \cite{Rotemberg2021}. The ISIC2020 dataset contains 33,126 images of 2,056 individual patients, with 32,542 benign and 584 malignant melanoma images. Although malignant melanoma and ugly duckling labels are not interchangeable as mentioned previously, we consider the melanoma labels as ugly ducklings with the ISIC2020 dataset for the purpose of evaluation. In addition, malignant melanoma being rare incidences, ISIC2020 also has a large class imbalance (ratio 55.7:1). This allow us to test if our proposed DMT-Quadruplet network is capable of still classifying the minority class accurately.

\subsection{Experiment setup}

The quadruplet network that serves as the main backbone of our proposed method is trained with a CNN feature extractor. By default, this is a ResNet18 model where the last layer has been replaced by a fully-connected layer that provides a 128-dimensional embedding as the output. For the quadruplet network, Adam optimizer is adopted during training with a learning rate of $0.0001$. The default margin values for $\alpha$ and $\beta$ are $1.0$ and $1.5$ respectively. 
% For evaluations, we compare the performance of different backbone feature extractors including ResNet34 and VGG16. 

During classification, the original last layer of the backbone CNN is added back in to provide a 2-dimensional embedding. The classifier is also trained with an Adam optimizer, with a Cross Entropy loss and a learning rate of $0.0001$. We compare our classification results with ResNet18 as a baseline traditional classifier. We also test a few different backbone architectures including ResNet34, VGG16, and EfficientNetB0. 

All experiments were carried out with PyTorch on a NVIDIA Tesla V100 32GB GPU. Both quadruplet and classifier networks were trained with a batch size of 32 and until the best accuracy was obtained with the least disparity among train and validation sets. 

To perform comprehensive performance comparison of the final classification task, we use seven metrics, including sensitivity, specificity, total accuracy, ROC AUC, and macro-weighted precision, recall, and F1 score. 

\section{Results and Discussion} \label{sec:results}

\begin{figure}[t]
\centerline{\includegraphics[width=0.5\textwidth, scale=0.5]{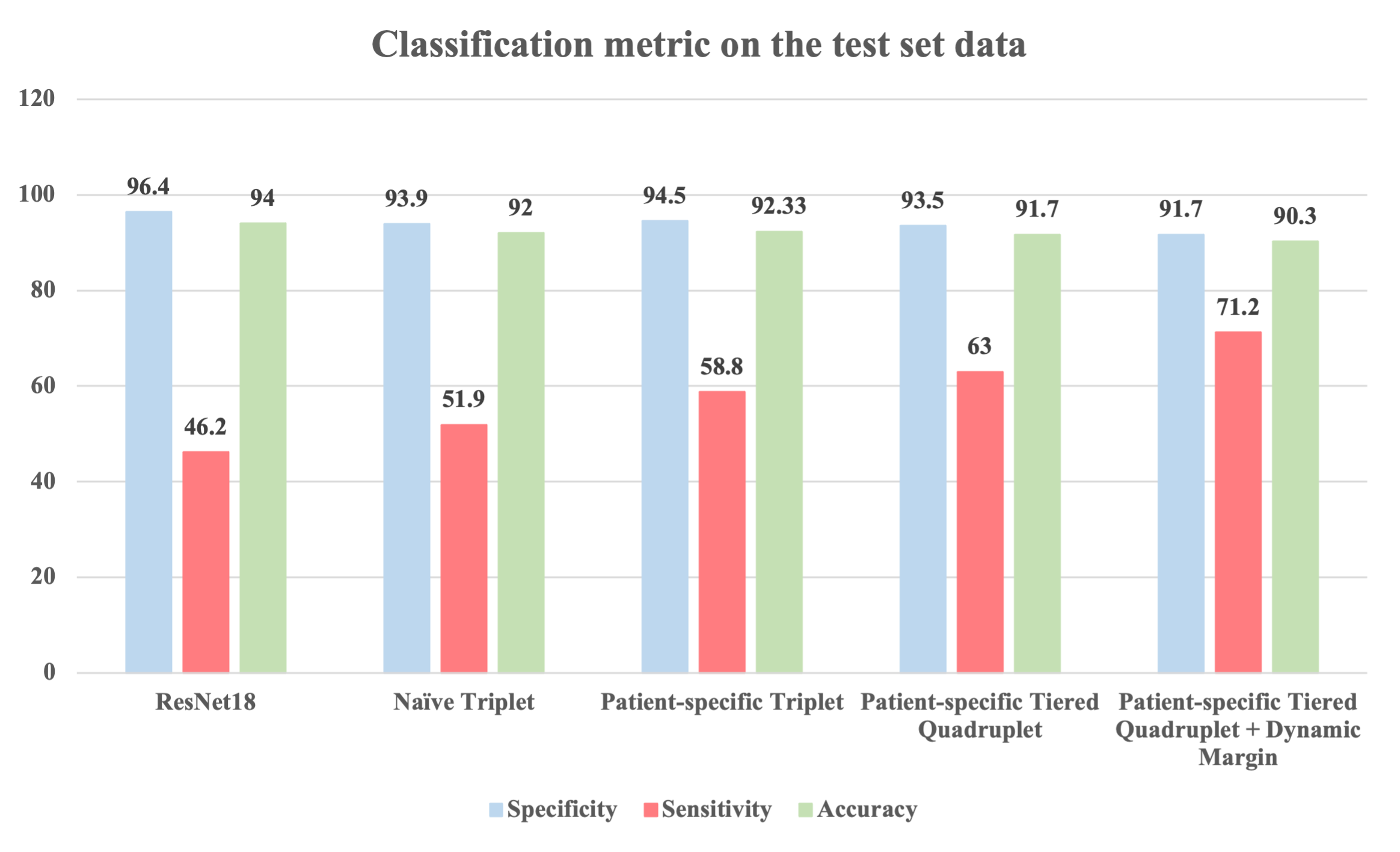}}
\caption{Performance of the triplet-based classifiers compared to the baseline. The sensitivity measure (classification accuracy on UDs) significantly increases with improvements to the triplet network, and the final DMT-Quadruplet network performs best with a reduced bias.}
\label{fig:4}
\end{figure}

\begin{table}[t]
\caption{Classification performance on the test set with different backbone CNNs of DMT-Quadruplet}\label{tab:table2}
\resizebox{\columnwidth}{!}{%
\begin{tabular}{|l|ccccccc|}
\hline
\textbf{Backbone models} & \textbf{Specificity} & \textbf{Sensitivity} & \textbf{Recall} & \textbf{Precision} & \textbf{F1-score} & \textbf{AUC} & \textbf{Accuracy} \\ \hline
ResNet18 & 91.7±1.4 & \textbf{71.2±5.9} & \textbf{81.4±2.2} & 62.2±0.8 & 66.3±0.9 & 90.2±0.0 & 90.3±1.2 \\
ResNet34 & 91.0±2.5 & 65.5±5.7 & 78.3±1.8 & 60.9±1.5 & 64.3±2.2 & 86.7±1.1 & 89.3±2.4 \\
VGG16 & 88.7±0.4 & 66.9±2.6 & 77.8±1.1 & 58.8±0.2 & 61.5±0.4 & 89.2±1.2 & 87.7±0.5 \\
EfficientNetB0 & \textbf{95.9±1.2} & 58.6±8.4 & 77.2±3.6 & \textbf{67.6±1.9} & \textbf{70.8±0.4} & \textbf{91.0±1.3} & \textbf{94.0±0.8} \\
DenseNet169 & 90.7±3.0 & 64.8±9.5 & 77.8±3.4 & 60.7±2.0 & 63.8±2.3 & 87.7±2.0 & 89.3±2.6 \\ \hline
\end{tabular}%
}
\end{table}

\begin{table}[t]
\caption{Classification performance on the test set with different classifiers of DMT-Quadruplet}\label{tab:table3}
\resizebox{\columnwidth}{!}{%
\begin{tabular}{|l|ccccccc|}
\hline
Classifier for Stage 2 & \multicolumn{1}{l}{Specificity} & \multicolumn{1}{l}{Sensitivity} & \multicolumn{1}{l}{Recall} & \multicolumn{1}{l}{Precision} & \multicolumn{1}{l}{F1-score} & \multicolumn{1}{l}{AUC} & \multicolumn{1}{l|}{Accuracy} \\ \hline
CNN & \textbf{91.6±0.2} & 73.8±1.8 & 82.6±0.9 & \textbf{62.3±0.2} & \textbf{66.6±0.3} & \textbf{91.6±0.3} & \textbf{90.3±0.5} \\
Logistic Regression & 89.4±0.3 & 77.0±0.7 & 83.2±0.2 & 60.7±0.2 & 64.3±0.3 & 90.1±0.4 & 88.9±0.3 \\
Random Forest & 89.3±0.5 & \textbf{78.4±0.8} & \textbf{83.7±0.4} & 60.5±0.2 & 64.0±0.3 & 90.3±0.3 & 88.5±0.1 \\
SVM & 90.0±0.3 & 77.3±1.4 & 83.6±0.8 & 61.2±0.4 & 65.1±0.7 & 90.2±0.6 & 89.5±0.4 \\ \hline
\end{tabular}%
}
\end{table}

\begin{table*}[t]
\caption{Performance comparison on the ISIC2020 dataset for binary classification}\label{tab:table4}
\resizebox{\textwidth}{!}{%
\begin{tabular}{|c|l|c|ccc|}
\hline
\textbf{Reference} & \multicolumn{1}{c|}{\textbf{Method}} & \textbf{Model} & \textbf{Sen} & \textbf{Spe} & \textbf{Acc} \\ \hline
\cite{Shah2020} & SMOTE-Tomek Links sampling for   data imbalance + Transfer learning & ResNet50 & 99.7 & 55.67 & 93.96 \\
\cite{Wan2023} & Long attention + Multi-scale   learning & MSLANet & 59.6 & 97.4 & 95.6 \\
\cite{Dong2023} & Semi supervised color mapping +   multi-scale attention mechanism & SSGNet & 64.1 & 94.67 & 89.68 \\
\cite{Arani2023} & WGAN for data imbalance + mobile   deep learning & Melanlysis & 92.5 & 89.5 & 94 \\
 & Our proposed method (patient-specific tiered quadruplet with dynamic margin) & DMT-Quadruplet & 71.4 & 73.3 & 72.7 \\ \hline
\end{tabular}%
}
\end{table*}

\subsection{Effectiveness of patient-specific Tiered Quadruplet approach}

This section evaluates the effectiveness of the proposed approach on the SkinUD dataset. We mainly focus on comparing the sensitivity measure as that is the metric that corresponds to the classification accuracy on UD lesions. Table \ref{tab:table1} shows that the Patient-specific Tiered Quadruplet with Dynamic Margin (DMT-Quadruplet) has the highest sensitivity measure for the SkinUD dataset at 71.2\% accuracy. This is 54\% better than the baseline ResNet18 classifier, while being 37\% better than the naive triplet classifier. 

Further, as illustrated in Figure \ref{fig:4}, all triplet-based classifiers consecutively improve upon the sensitivity measure, all reporting higher accuracy on classifying UD lesions than the baseline. This causes a trade-off between specificity and sensitivity as the bias caused by the majority class is reduced, resulting in specificity and overall accuracy declining slightly. However, the specificity and overall accuracy of the DMT-Quadruplet is lower than the baseline ResNet18 only by 4.8\% and 3.9\% respectively. Moreover, the DMT-Quadruplet reports the highest recall and ROC AUC as well.

In addition, it can be seen that both Tiered Quadruplet and DMT-Quadruplet outperforms the patient-specific triplet in sensitivity. This further highlights the importance of our Tiered Quadruplet loss as it is capable of incorporating more lesion-level information with the help of the secondary-negative, which provides the network with an additional global view of lesion features. 

Overall, coupled with the patient-specific mining approach, computing an online dynamic margin of separation boosts the performance of the DMT-Quadruplet network. This could be due to the fact that having a fixed margin earlier would have inaccurately discarded some of the useful triplets from individuals during the online mining process. Having a patient-specific online dynamic margin helps the mining process to pick more accurate and useful triplets from each individual, feeding better representations to the final loss of each iteration. Further, by setting a large value for the patient-level separating margin $\beta$ of the DMT-Quadruplet, more weight is applied on learning global features than the finer local features during training. This allows the network to remain patient-agnostic at lesion-level while separating lesions at patient-level accurately.

\begin{figure}[!t]
\centering
\begin{subfigure}{.25\textwidth}
  \centering
  \includegraphics[width=1\linewidth]{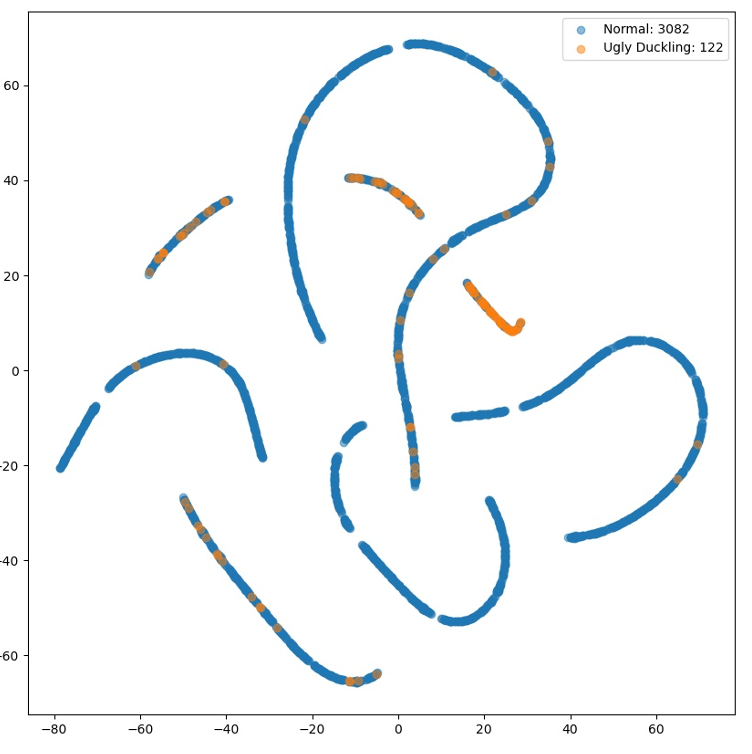}
  \caption{For the Naive triplet}
  \label{fig:sub1}
\end{subfigure}%
\begin{subfigure}{.25\textwidth}
  \centering
  \includegraphics[width=1\linewidth]{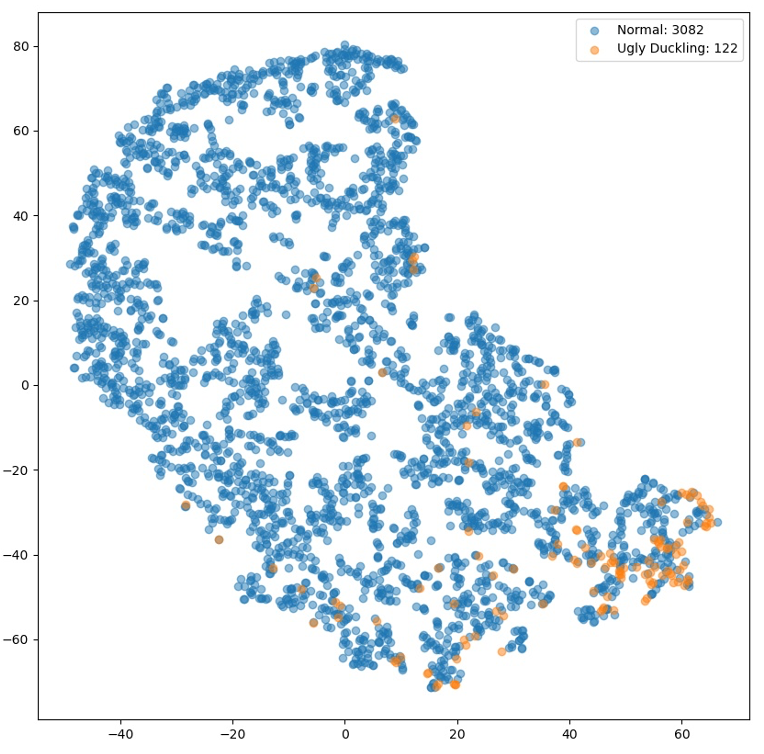}
  \caption{For the DMT-Quadruplet}
  \label{fig:sub2}
\end{subfigure}
\caption{The 2D separation manifold in metric space for all samples in the test set using (a) the Naive triplet and (b) the DMT-Quadruplet to generate embeddings. The 2D embeddings were generated by applying t-SNE algorithm on the 128-dimensional output of the triplet networks.}
\label{fig:10}
\end{figure}

The effectiveness of the DMT-Quadruplet approach is further proven when we visualize the separation of lesions at both patient-level and lesion-level. We applied t-SNE algorithm to obtain 2-d embeddings from the 128-dimensional embeddings generated for samples in the test set to illustrate the cluster separations as shown in Figure ~\ref{fig:10}. The manifold of the Naive triplet does not show a good separation for UD lesions (Figure ~\ref{fig:10}(a)) with UDs spread over the manifold, but the DMT-Quadruplet shows a better separation by clustering UD lesions closer together (Figure ~\ref{fig:10}(b)).

\begin{figure}[!t]
\centering
\begin{subfigure}{.25\textwidth}
  \centering
  \includegraphics[width=1\linewidth]{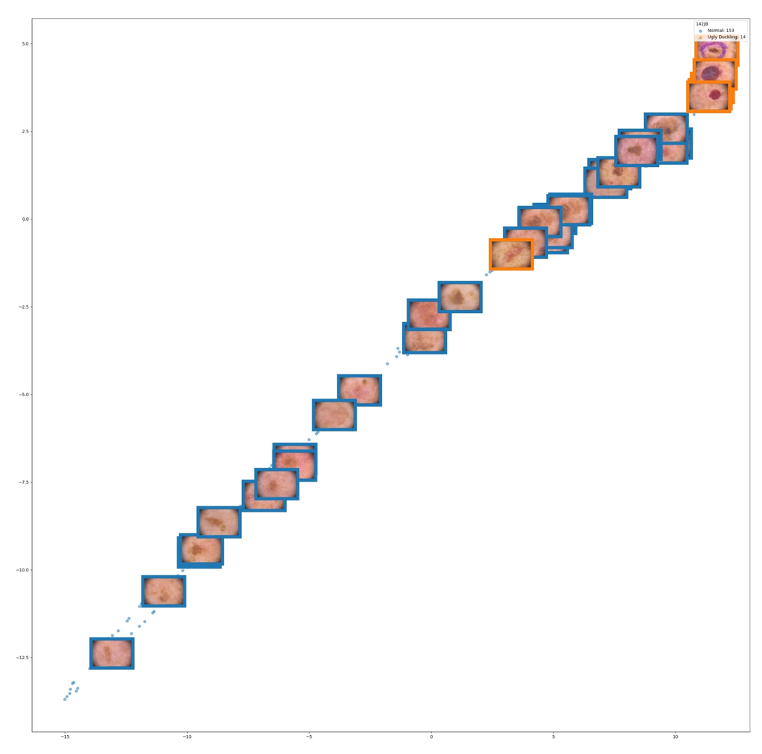}
  \caption{For the Naive triplet}
  \label{fig:sub3}
\end{subfigure}%
\begin{subfigure}{.25\textwidth}
  \centering
  \includegraphics[width=1\linewidth]{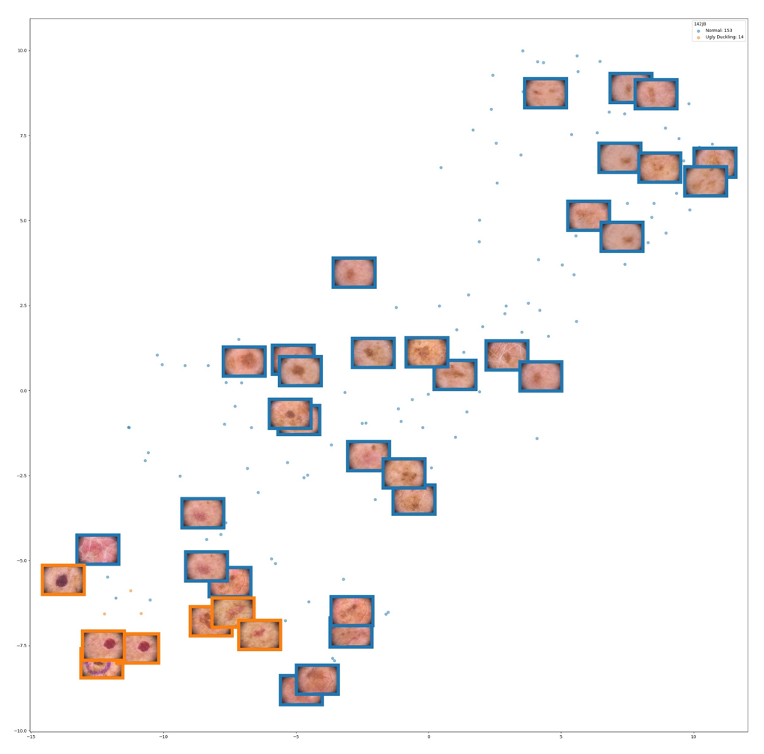}
  \caption{For the DMT-Quadruplet}
  \label{fig:sub4}
\end{subfigure}
\caption{The 2D separation manifold with image tiles in metric space for the same individual in the test set using (a) the Naive triplet and (b) the DMT-Quadruplet to generate embeddings. For ease of interpretation, not all image tiles for normal lesions have been plotted. However, all UD lesion tiles are plotted.}
\label{fig:11}
\end{figure}

Additionally, as shown in Figure \ref{fig:11}, the two networks behave differently when differentiating between all lesions of the same patient, where the DMT-Quadruplet shows a clearer clustering. Thus, even though the mining of triplets for the DMT-Quadruplet was patient-specific, it is capable of behaving in a patient-agnostic manner while separating lesions within an individual equally well.

\subsection{Analyses on different feature extraction backbones for DMT-Quadruplet}

Table \ref{tab:table2} presents the classification performance of the DMT-Quadruplet network with different backbone CNNs. The results indicate that EfficientNetB0 performs better at classifying the majority class with a higher sensitivity and overall accuracy. However, it leads to a lower sensitivity measure, indicating poor classification of the minority (UD) class. ResNet18 reports the highest sensitivity, while maintaining high specificity, AUC and overall accuracy. Thus, ResNet18 can be considered as the most suitable feature extractor for DMT-Quadruplet. 

\subsection{Analyses on different classifiers for classification of DMT-Quadruplet Embeddings}

For the embeddings generated by the trained DMT-Quadruplet, we tested different classifiers for our stage 2 of the proposed architecture. As presented in Table \ref{tab:table3}, the results indicate that a Random Forest classifier with a depth of 4 is performs best in classifying UDs (sensitivity 78.4\%) with the highest recall (83.7\%). However, our original CNN classifier remains the best performing for all other metrics.

% \subsection{Analyses on different skin lesion datasets}

\subsection{Classification on ISIC2020 dataset}

As patient-specific data is available in ISIC2020 dataset, we trained our DMT-Quadruplet on it separately. Evaluation on the test set indicates that the trained model can achieve more than 70\% accuracy on both benign and melanoma classes. 

We present the comparison of the performance of our method with existing work in Table \ref{tab:table4}. For this comparison, we only considered work where binary classification was performed on the ISIC2020 dataset. Further, we omitted methods using a combination of ISIC datasets, methods with ensemble models, and methods not reporting sensitivity and specificity for a fair comparison. While \cite{Arani2023} reported the highest sensitivity and specificity among the considered work, it should be noted that they address the issue of class imbalance in ISIC2020 by generating synthetic skin lesion images using an improved Generative Adversarial Network (GAN). This results in a more balanced dataset, and better classification accuracy. However, our method only uses duplication of the minority class to address the class imbalance, yet achieving $\geq$70\% in accuracy for both classes with DMT-Quadruplet. Further, the results of \cite{Wan2023} and \cite{Dong2023} show that there's a high trade-off between sensitivity and specificity. While both these works obtained a high specificity, their sensitivity was considerably low, indicating poorer performance in classifying the melanoma lesions. However, our proposed DMT-Quadruplet method achieves similar accuracy for both classes. Thus, if we applied more advanced data augmentation techniques to address the class imbalance in ISIC2020, DMT-Quadruplet would further improve in performance. Altogether, the results indicate that our proposed DMT-Quadruplet model is generalisable across skin lesion datasets with patient-specific metadata.

Although previous work on metric learning have implemented similar hierarchical triplet networks with a layered ontology, the datasets that have been used in training these models had mutually exclusive sub-classes (eg: types of shoes and types of trousers in the Fashion60 dataset) \cite{He2021}. However, our problem is unique in the sense that it has a mutual ontology where the sub-classes (lesion classes within an individual) are the same as the classes of the overall dataset. For the whole dataset, there are only two sub-classes of lesions (normal and UD), although they differ slightly from one individual to another. Thus, our experimental evaluation indicates that our DMT-Quadruplet loss captures inclusive intra-class and inter-class relationships optimally and accurately.

\subsection{Limitations}

It must be noted that the manual assessment and annotation of ugly duckling lesions are subjective to inter-observer variability. Based on expertise and skill, dermatologists may vary among themselves in what they identify as an ugly duckling lesion among normal lesions. For our study, we considered annotations by only one board-certified dermatologist. For more non-biased ground truth annotations, future work could improve by considering a majority vote on ugly ducklings from several dermatologists.

Further, our SkinUD dataset is comprised of an entirely Caucasian population. Thus, our work may lack generalisability outside Caucasian individuals as populations with skin of colour may present ugly duckling features differently. In people with skin of colour, a highly pigmented lesion might not always show signs of an ugly duckling. Thus, our work could benefit from incorporating data from skin of colour. However, currently there are no public skin lesion datasets that include non-Caucasian individuals, nor for annotations of ugly duckling lesions.

In addition, due to lack of data and manual annotation resources, our SkinUD dataset is considerably small with only 37 participants. A larger sample size with annotations would help improve any future work.

\section{Conclusion}

In this work, we present a a novel patient-specific metric learning method for improved classification of ugly duckling lesions. For this, we implement a quadruplet sampling strategy that enables the network to learn features from two tiers of information, where tier 1 is lesion features between individuals, and tier 2 is lesion features within an individual. Considering our problem is unique with mutual sub-classes between tier 1 individuals, we introduce a tiered quadruplet network to incorporate more global context for network training. We further improve our tiered quadruplet network by including an inter-patient dynamic margin of separation to drive the network to mine for tailored and useful triplets from individuals.

The effectiveness of our proposed approach is demonstrated by the extensive experimental results. Our Tiered Quadruplet network with a Dynamic margin (DMT-Quadruplet) is shown to effectively capture global lesion-level features, while learning finer, patient-level differences as well. 

In addition, our optimized approach surpasses traditional classification methods for identifying lesions of the minority class, while also being capable of handling datasets with major class imbalances. As future work, a class-center based triplet loss can be incorporated to the tiered quadruplet network to validate if the class imbalance can be further alleviated. 

In conclusion, with our two-stage pipeline combining a triplet-based feature extractor and a classifier, we can effectively separate lesions of an individual and accurately classify ugly duckling lesions. In clinical application, our method will be particularly useful for patients who have many naevi ($>$100) as assessing each lesion individually is time consuming and unrealistic for a clinician. Applying this alongside an ABCDE algorithm might provide an image triage so the dermatologist/clinician only needs to assess the most suspicious lesions, saving time and effort. As such, the proposed method can successfully assist clinicians in early melanoma detection as ugly duckling lesions are an indicator of a potential malignant melanoma developing.

\section{Conflicts of interest}\label{COI}
HPS is a shareholder of MoleMap NZ Limited and e-derm consult GmbH and undertakes regular teledermatological reporting for both companies. HPS is a Medical Consultant for Canfield Scientific Inc, Blaze Bioscience Inc, and a Medical Advisor for First Derm.

\section{Author Contributions}\label{contributions}
\textbf{Nathasha Naranpanawa:} Conceptualization, Data Curation, Methodology, Software, Validation, Formal analysis, Writing - Original Draft, Writing - Review \& Editing,  Visualization
\textbf{H. Peter Soyer:} Conceptualization, Data Curation, Resources, Writing - Review \& Editing
\textbf{Adam Mothershaw:} Software, Resources, Writing - Review \& Editing
\textbf{Gayan K. Kulatilleke:} Writing - Review \& Editing
\textbf{Zongyuan Ge:} Writing - Review \& Editing
\textbf{Brigid Betz-Stablein:} Conceptualization, Data Curation, Writing - Review \& Editing, Supervision
\textbf{Shekhar S. Chandra:} Conceptualization, Methodology, Writing - Review \& Editing, Supervision

\section{Funding}\label{sec:funding}
The images used in this study are derived from the BNMS – Brisbane Naevus Morphology Study. This BNM Study was approved by the Metro South Human Research Ethics Committee (approval \#HREC/09/QPAH/ 162, 26 August 2009) and The University of Queensland (approval \#2009001590, 14 October 2009) and conducted in accordance with the Declaration of Helsinki. Participants of the BNM Study provided written consent after receiving a Participant Information and Consent Form. The BNM Study was funded by the Australian National Health and Medical Research Council (NHMRC) (project grants APP1062935, APP1083612) and the Centre of Research Excellence for the Study of Naevi (grant no. APP1099021). 

The above mentioned funding sources of the BNM Study had no involvement in the work presented in this paper.

\bibliographystyle{IEEEtran}
\bibliography{bibliography}

\end{document}